\documentclass{article}

\usepackage{multirow} 
\usepackage{graphicx}
\usepackage{amsmath}
\usepackage{algorithm}
\usepackage{booktabs}
\usepackage{subfigure}
\usepackage{wrapfig}
\usepackage{booktabs}

\usepackage{pdfpages}

\usepackage{amsmath,amsfonts,bm}

\usepackage{amssymb,amsthm,mathtools}

\def\Figref#1{Figure~\ref{#1}}

\def\eqref#1{equation~\ref{#1}}

\def\1{\bm{1}}

\DeclareMathAlphabet{\mathsfit}{\encodingdefault}{\sfdefault}{m}{sl}
\SetMathAlphabet{\mathsfit}{bold}{\encodingdefault}{\sfdefault}{bx}{n}

 \usepackage[preprint]{neurips_2025}

\usepackage[utf8]{inputenc} 
\usepackage[T1]{fontenc}    
\usepackage{hyperref}       
\usepackage{url}            
\usepackage{booktabs}       
\usepackage{amsfonts}       
\usepackage{nicefrac}       
\usepackage{microtype}      
\usepackage{xcolor}         

\title{Purified OPSD: On-Policy Self-Distillation Without Losing How to Think}

\author{
  \mdseries
  Zhanming Shen$^{1,2}$\quad
  Jintao Tong$^{2,3}$\quad
  Shaotian Yan$^{2}$\quad
  Chen Shen$^{2\dagger\ddagger}$\quad
  Hao Chen$^{1,2}$\\
  Wentao Ye$^{1}$\quad
  Xiaomeng Hu$^{1,2}$\quad
  Rui Miao$^{2,4}$\quad
  Haobo Wang$^{1\dagger}$\quad
  Junbo Zhao$^{1}$\\
  Gang Chen$^{1}$\quad
  Jieping Ye$^{2}$\\[6pt]
  $^{1}$Zhejiang University\quad
  $^{2}$Tongyi Lab, Alibaba Group\\
  $^{3}$Huazhong University of Science and Technology\quad
  $^{4}$Jilin University\\[3pt]
  {\small $^\dagger$Corresponding Author\quad $^\ddagger$Project Leader}
}

\begin{document}

\maketitle

\begin{abstract}
On-policy self-distillation (OPSD) has emerged as a promising paradigm for improving LLM reasoning, where a privileged teacher with access to reference solutions provides token-level supervision on the student's own generated trajectories. However, we find that OPSD consistently fails on long chain-of-thought (long-CoT) reasoning models, yielding at best marginal gains while destabilizing the reflective reasoning capability these models depend on. Through a novel decomposition of the teacher's supervision signal, we identify the root cause: the teacher's supervision is dominated by a reference-induced component that drives rote memorization of reference-specific shortcuts, while the question-conditioned, inference-transferable component is ignored or actively opposed. Based on this diagnosis, we propose a two-step solution. First, we construct a \emph{reference-only teacher}, i.e., the same model conditioned on the reference without the question, to isolate the non-transferable component of the supervision signal; the residual after subtracting this component captures the question-conditioned, inference-transferable correction. Second, we use pointwise mutual information (PMI) as the mechanism to transform this residual into a well-formed \emph{PMI target} distribution that the student can directly distill from, filtering out the reference-induced shortcut. Experiments on four long-CoT models across two datasets demonstrate consistent improvements over both the base model and standard OPSD, while preserving the models' natural epistemic behavior throughout training.
\end{abstract}

\section{Introduction}
\label{sec:intro}

Recent advances in large language model (LLM) reasoning have demonstrated that long chain-of-thought (long-CoT) reasoning, where models engage in extended, reflective thinking processes before arriving at answers, can substantially improve performance on complex reasoning tasks~\citep{jaech2024openai,guo2025deepseek,team2024qwq}. Distilling these capabilities into smaller, more efficient models has become a central challenge, with on-policy self-distillation (OPSD)~\citep{agarwal2024onpolicy,he2026self,zhao2026self,cui2026brief} emerging as a promising paradigm. In OPSD, a student model generates reasoning trajectories on-policy, which are then evaluated by a privileged teacher that has access to reference solutions, providing token-level supervision through divergence minimization.

\begin{figure}[h]
\centering
\subfigure[Qwen3-8B]{
    \includegraphics[width=0.23\textwidth,keepaspectratio]{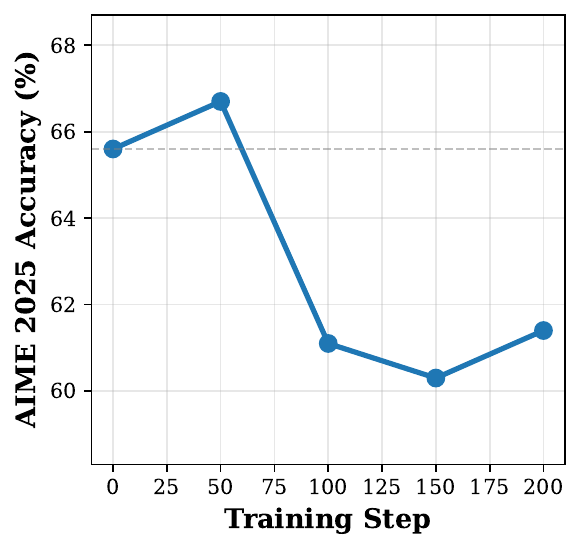}
    \label{fig:collapse-8b}
}
\subfigure[Qwen3-4B]{
    \includegraphics[width=0.23\textwidth,keepaspectratio]{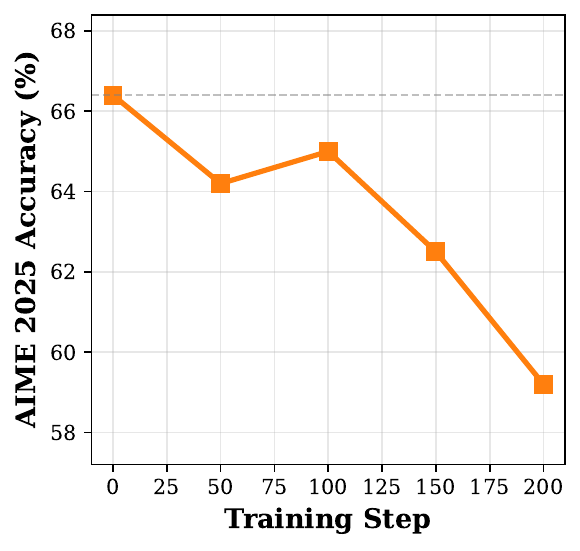}
    \label{fig:collapse-4b}
}
\subfigure[R1-Distill-7B]{
    \includegraphics[width=0.23\textwidth,keepaspectratio]{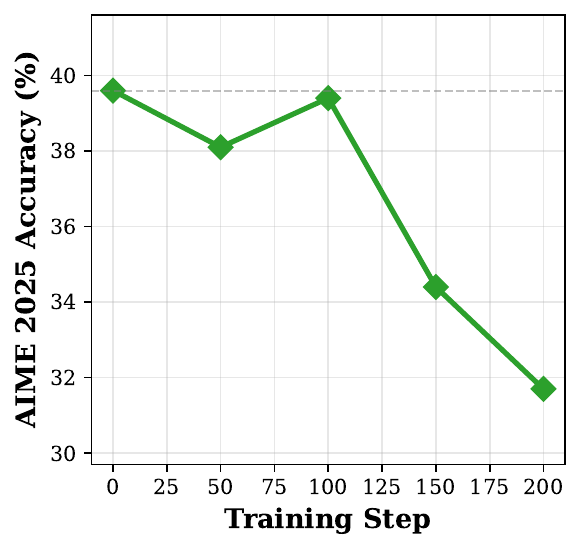}
    \label{fig:collapse-r1}
}
\subfigure[OLMo-7B]{
    \includegraphics[width=0.23\textwidth,keepaspectratio]{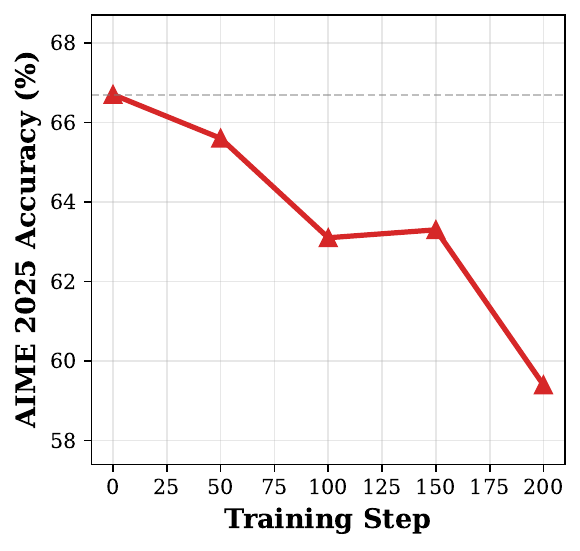}
    \label{fig:collapse-olmo}
}
\caption{AIME 2025 accuracy across OPSD training checkpoints on Math-CoT-20K. OPSD provides at best marginal, short-lived gains, and may even degrade performance on long-CoT models. This finding is corroborated by concurrent works~\citep{kaur2026rethinking,kim2026does}, which report similarly limited or negative outcomes when applying OPSD to thinking models.}
\label{fig:opsd-collapse}
\end{figure}

However, \textbf{applying OPSD to long-CoT reasoning models yields surprisingly poor results}. As shown in \Figref{fig:opsd-collapse}, our systematic evaluation across four long-CoT models reveals that OPSD provides at best marginal, short-lived gains, and may even degrade performance on long-CoT models. This finding is corroborated by concurrent works~\citep{kaur2026rethinking,kim2026does}, which report similarly limited or negative outcomes when applying OPSD to thinking models.

\textbf{What goes wrong?} To investigate, we examine the model's generated outputs across training checkpoints. Following \citet{kim2026does,kim2026understanding}, we track epistemic markers (tokens such as ``\textit{Wait}'', ``\textit{Let me think}'', ``\textit{Perhaps}'', and ``\textit{Maybe}'') that indicate uncertainty externalization and reflective reasoning, which are central to long-CoT models' generalization ability~\citep{gandhi2024stream,li2025feature}. We observe a striking and pathological pattern (\Figref{fig:epistemic-markers}): as OPSD training progresses, these epistemic markers exhibit \emph{violent, erratic fluctuations} across different models. On Qwen3-8B, the total count collapses; on R1-Distill-Qwen-7B, it explodes with the increase concentrated almost entirely on the single token ``\textit{Wait}''. When combined with the steady performance degradation, this pattern suggests that OPSD's answer-directed teaching paradigm \textbf{may not stably preserve the reflective reasoning capability} that long-CoT models fundamentally depend on for generalization.

This observation leads us to a natural hypothesis: OPSD's privileged teacher, having access to the reference solution, already knows the answer and therefore has no need to reflect, explore, or self-correct. Rather than providing selective, beneficial corrections to the student's reasoning (the intended purpose of privileged supervision), \textbf{the teacher may instead be driving the student toward rote memorization of the specific reasoning path dictated by the reference, destabilizing the model's ability to think independently.}

We verify this hypothesis through a novel decomposition of the teacher's supervision signal. We construct a reference-only teacher (the same model conditioned on the reference solution without the question), and use it to decompose the teacher's supervision into two components: a reference-induced component (the supervision that exists even without the question) and the remaining question-conditioned component. Our analysis across training checkpoints reveals a striking finding: \textbf{the reference-induced component dominates both the direction and magnitude of the teacher's update} (\Figref{fig:signal-dominance}), while the question-conditioned component, which genuinely helps solve the problem, remains nearly orthogonal or even anti-aligned with the total update direction, and, especially in the early training phase, accounts for only a small fraction of the update norm. These findings directly corroborate the performance and reflection destabilization, forming a coherent explanation: OPSD collapses on long-CoT models because it drives rote memorization of reference-specific shortcuts rather than transferable reasoning improvement.

\textbf{A natural solution emerges from the diagnosis.} Our decomposition reveals that the reference-only teacher serves as a precise probe for identifying the non-transferable component of the supervision signal. Once this component is identified, the \emph{\textbf{residual}}, obtained by subtracting the reference-induced signal from the teacher's total update, can be interpreted as a conditional PMI-style quantity that captures the question-conditioned, inference-transferable correction we want the student to learn. However, this residual exists only as a per-token log-probability difference, not as a distribution the student can directly distill from. We leverage pointwise mutual information (PMI), a classical tool from information theory widely used in NLP for measuring association strength~\citep{church1990word,levy2014neural}, to transform this residual into a well-formed target distribution.

Concretely, standard OPSD distills from the raw teacher distribution $\pi_\mathrm{T}$, which we have shown to be dominated by the reference-induced shortcut:
\begin{equation}
    \mathcal{L}_{\mathrm{OPSD}} = D_{\mathrm{JSD}}\!\left(\pi_\theta \;\|\; \pi_\mathrm{T}\right).
\end{equation}
We replace $\pi_\mathrm{T}$ with a \textbf{purified PMI target} that strips out the reference-induced shortcut using the reference-only teacher $\pi_\mathrm{ref}$:
\begin{equation}
    \mathcal{L}_{\mathrm{ours}} = D_{\mathrm{JSD}}\!\left(\pi_\theta \;\|\; P_{\mathrm{PMI}}\right), \quad P_{\mathrm{PMI}}(v) \propto P_0(v) \exp\!\left(\tfrac{1}{\beta}\bigl(\log \pi_\mathrm{T}(v) - \log \pi_\mathrm{ref}(v)\bigr)\right),
\end{equation}
where $P_0$ is the clean base model distribution (without reference), $\pi_\mathrm{ref}$ is the reference-only teacher that isolates the non-transferable component, and $\beta$ controls the correction strength. The difference $\log \pi_\mathrm{T} - \log \pi_\mathrm{ref}$ removes the reference-specific shortcut from the teacher's prediction, leaving only the question-conditioned correction. This target can be seamlessly integrated into the standard OPSD framework, achieving results far superior to standard OPSD and protecting the natural epistemic behavior of the base model.


\section{Revisiting On-Policy Self-Distillation for Long-CoT Models}
\label{sec:revisit}

On-Policy Self-Distillation (OPSD)~\citep{agarwal2024onpolicy,he2026self,zhao2026self,cui2026brief} has shown promise for improving reasoning capabilities through privileged teacher guidance. However, when applied to long chain-of-thought (long-CoT) reasoning models, i.e., models that rely on extended, reflective reasoning processes, we observe a surprising and consistent failure. In this section, we first introduce the relevant background, then systematically investigate this failure through three progressive analyses.

\subsection{Preliminaries}
\label{sec:prelim}

\paragraph{On-Policy Self-Distillation.}
OPSD trains a student model $\pi_\theta$ to learn from a privileged teacher that shares the same base model $\pi_0$ but has access to a reference solution $r$. The student generates reasoning trajectories on-policy: $\hat{y} \sim \pi_\theta(\cdot \mid q)$, and the teacher evaluates them token-by-token. The training objective minimizes:
\begin{equation}
    \mathcal{L}_{\mathrm{OPSD}} = \frac{1}{T} \sum_{t=1}^{T} D_{\mathrm{JSD}}\!\left(\pi_\theta(\cdot \mid \hat{y}_{<t}, q) \;\|\; \pi_\mathrm{T}(\cdot \mid \hat{y}_{<t}, q, r)\right),
    \label{eq:opsd}
\end{equation}
where $D_{\mathrm{JSD}}$ denotes the generalized Jensen-Shannon divergence.

\paragraph{Pointwise Mutual Information.}
Pointwise mutual information (PMI) is a classical measure from information theory that quantifies the association between two events beyond what would be expected under independence~\citep{church1990word,bouma2009normalized}. Given two random variables $X$ and $Y$, the PMI of a specific outcome pair $(x, y)$ is $\mathrm{PMI}(x; y) = \log \frac{P(x, y)}{P(x) P(y)}$. PMI has been widely used in NLP for collocation detection~\citep{church1990word}, word embedding evaluation~\citep{levy2014neural}, and more recently for analyzing the information content of language model predictions~\citep{holtzman2021surface}. In our setting, we adopt a PMI-style conditional log-likelihood ratio to measure how much the teacher's prediction changes when the question is introduced on top of the reference-conditioned context, enabling us to separate inference-transferable supervision from reference-induced shortcuts (Section~\ref{sec:revisit-diagnosis}).

\subsection{OPSD Degrades Long-CoT Model Performance}
\label{sec:revisit-fail}

We apply standard OPSD to four long-CoT reasoning models (Qwen3-8B, Qwen3-4B, DeepSeek-R1-Distill-Qwen-7B, and OLMo-7B-Thinking) using Math-CoT-20K with reference solutions as privileged information. We evaluate performance across training checkpoints on the AIME 2025 benchmark.

As shown in \Figref{fig:opsd-collapse}, OPSD provides at best marginal, short-lived gains, and may even degrade performance on long-CoT models. This finding is corroborated by concurrent works~\citep{kaur2026rethinking,kim2026does}, which report similarly limited or negative outcomes when applying OPSD to thinking models.

\subsection{The Destabilization of Reflective Reasoning}
\label{sec:revisit-reflection}

To understand the mechanism behind performance degradation, we examine the model's generated outputs across training checkpoints on the AIME test set. Following \citet{kim2026does}, we track epistemic markers across OPSD training checkpoints. They indicate uncertainty externalization and reflective reasoning, which are central to long-CoT models' generalization ability~\citep{gandhi2024stream,li2025feature}.

\begin{figure}[t]
\centering
\subfigure[Qwen3-8B: epistemic count]{
    \includegraphics[width=0.23\textwidth,keepaspectratio]{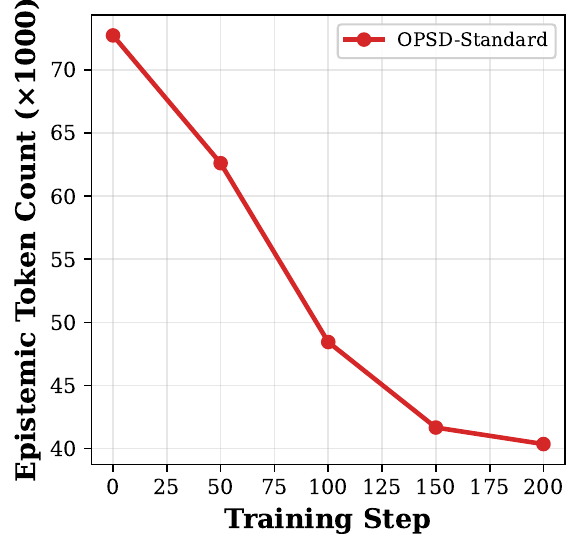}
    \label{fig:qwen-traj}
}
\subfigure[Qwen3-8B: marker distribution]{
    \includegraphics[width=0.23\textwidth,keepaspectratio]{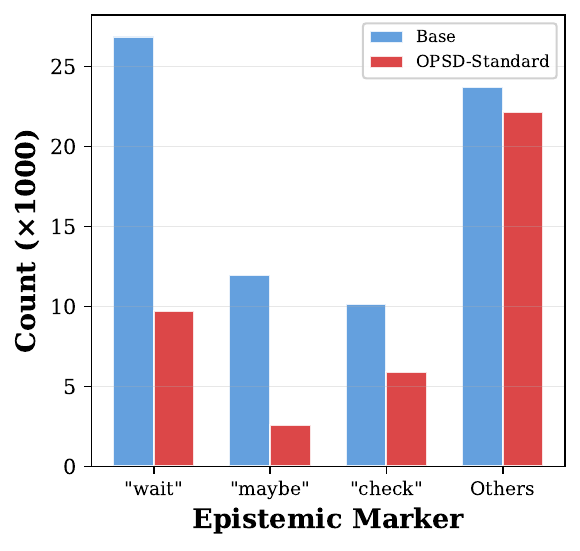}
    \label{fig:qwen-markers}
}
\subfigure[R1-7B: epistemic count]{
    \includegraphics[width=0.23\textwidth,keepaspectratio]{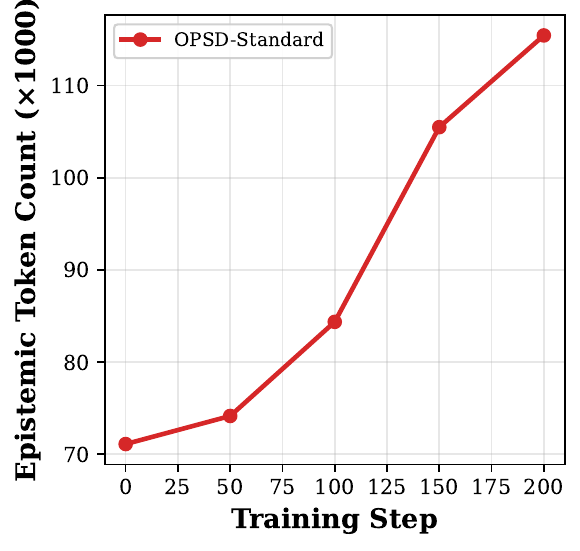}
    \label{fig:r1-traj}
}
\subfigure[R1-7B: marker distribution]{
    \includegraphics[width=0.23\textwidth,keepaspectratio]{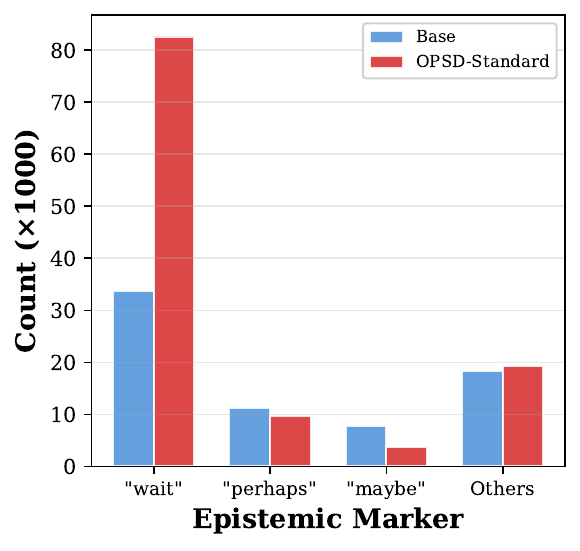}
    \label{fig:r1-markers}
}
\caption{Epistemic marker analysis during OPSD training on Math-CoT-20K. Qwen3-8B collapses uniformly, while R1-Distill-Qwen-7B explodes with the increase concentrated almost entirely on ``\textit{Wait}''. Both patterns are pathological and correlate with performance degradation.}
\label{fig:epistemic-markers}
\end{figure}

As shown in \Figref{fig:epistemic-markers}, we observe pathological behavior: On Qwen3-8B, epistemic markers collapse, with nearly all marker types declining uniformly; On R1-Distill-Qwen-7B, the total count explodes, but this increase is concentrated almost entirely on the single token ``\textit{Wait}'' (from 34K to 83K), while other markers decrease, suggesting degenerate repetition rather than genuine deliberation.

In both cases, the epistemic marker distribution undergoes violent shifts that correlate with performance degradation. This observation leads us to a natural hypothesis: OPSD's privileged teacher, having access to the reference solution, already knows the answer and therefore has no need to reflect, explore, or self-correct. Rather than providing selective, beneficial corrections to the student's reasoning (the intended purpose of privileged supervision), the teacher may instead be driving the student toward \textbf{rote memorization of the specific reasoning path dictated by the reference, destabilizing the model's ability to think independently.}

\subsection{Diagnosing the Root Cause: Rote Memorization of Privileged Information}
\label{sec:revisit-diagnosis}

The destabilization of reflective reasoning suggests that OPSD may be driving the student toward rote memorization of the specific reasoning path dictated by the reference, rather than learning transferable corrections. To test this, we decompose the teacher's supervision into interpretable components.

\paragraph{Decomposition of OPSD supervision.}
We construct a \emph{\textbf{reference-only teacher}} $\pi_\mathrm{ref}$, defined as the same model conditioned only on the reference solution $r$ without the question $q$. This enables us to decompose the teacher's total update relative to the current student $\pi_\theta$ into two components:
\begin{align}
    \underbrace{\log \pi_\mathrm{T} - \log \pi_\theta}_{\Delta_\mathrm{total}} &= \underbrace{\log \pi_\mathrm{ref} - \log \pi_\theta}_{\Delta_\mathrm{ref}:\;\text{reference-induced}} + \underbrace{\log \pi_\mathrm{T} - \log \pi_\mathrm{ref}}_{\Delta_\mathrm{it}:\;\text{inference-transferable}}, \label{eq:decomp}
\end{align}
where $\pi_\theta$ is the current student model at each checkpoint. The \textbf{reference-induced signal} $\Delta_\mathrm{ref}$ captures supervision driven purely by the reference solution, i.e., information the student will never have access to during inference. The remaining \textbf{inference-transferable signal} $\Delta_\mathrm{it}$ could capture \textbf{supervision that depends on understanding the problem and would remain useful at inference time when no reference is available.} If OPSD is working as intended, $\Delta_\mathrm{total}$ should align primarily with $\Delta_\mathrm{it}$; if it is memorizing the reference, it should align with $\Delta_\mathrm{ref}$.

\paragraph{Experimental setup.}
For each OPSD checkpoint, we generate on-policy trajectories using the current student $\pi_\theta$, then compute the decomposition above using the frozen base model for $\pi_\mathrm{T}$ and $\pi_\mathrm{ref}$. We measure two complementary metrics: (1)~\textbf{cosine similarity} $\cos(\Delta_\mathrm{total}, \Delta_\mathrm{ref})$ and $\cos(\Delta_\mathrm{total}, \Delta_\mathrm{it})$, capturing the \emph{directional} alignment; and (2)~\textbf{norm fraction} $\|\Delta_\mathrm{ref}\| / \|\Delta_\mathrm{total}\|$ and $\|\Delta_\mathrm{it}\| / \|\Delta_\mathrm{total}\|$, capturing the \emph{magnitude} dominance. Both are computed over the full vocabulary at each token position and averaged across 100 samples per checkpoint.

\begin{figure}[t]
\centering
\subfigure[Qwen3-8B: cosine similarity]{
    \includegraphics[width=0.23\textwidth,keepaspectratio]{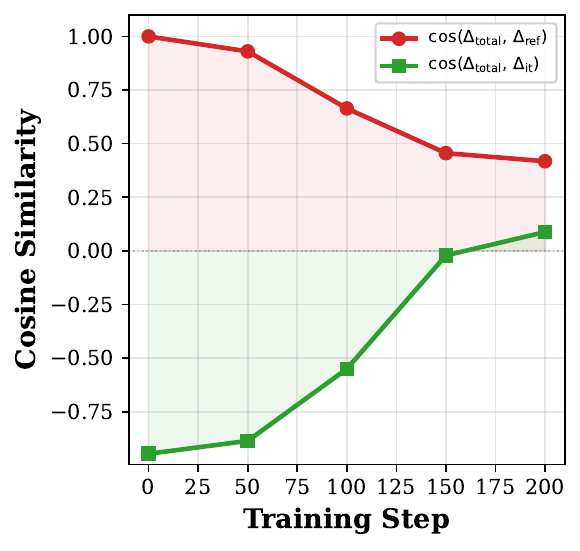}
    \label{fig:grad-align-8b}
}
\subfigure[Qwen3-8B: norm fraction]{
    \includegraphics[width=0.23\textwidth,keepaspectratio]{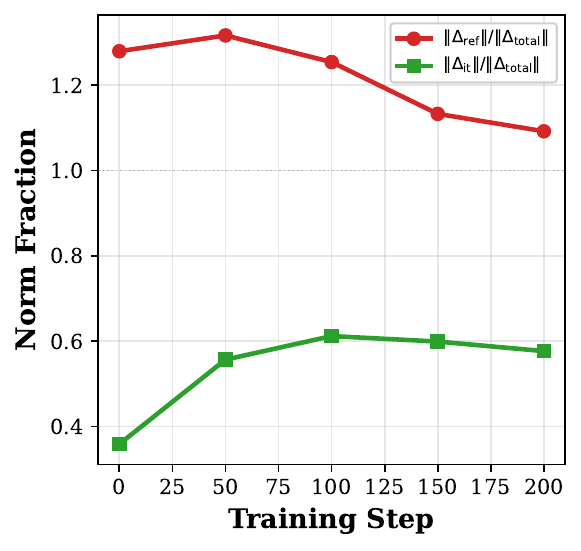}
    \label{fig:norm-frac-8b}
}
\subfigure[R1-7B: cosine similarity]{
    \includegraphics[width=0.23\textwidth,keepaspectratio]{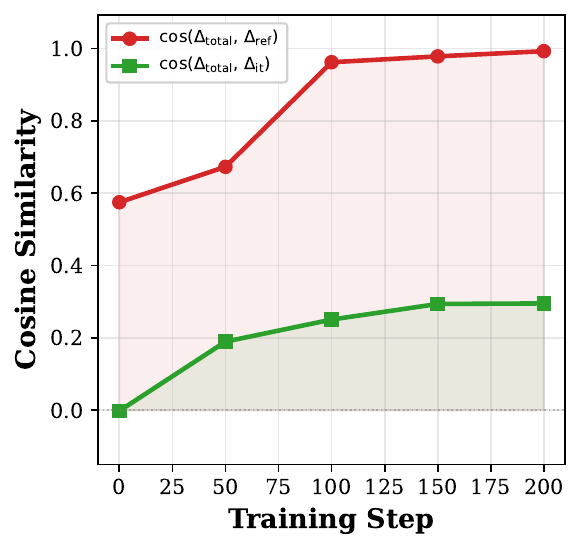}
    \label{fig:grad-align-r1}
}
\subfigure[R1-7B: norm fraction]{
    \includegraphics[width=0.23\textwidth,keepaspectratio]{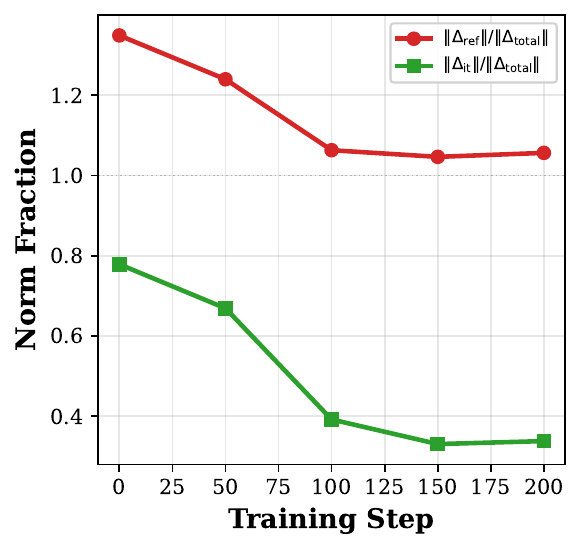}
    \label{fig:norm-frac-r1}
}
\caption{Decomposition of the teacher's update $\Delta_\mathrm{total} = \log \pi_\mathrm{T} - \log \pi_\theta$ into reference-induced ($\Delta_\mathrm{ref}$, red) and inference-transferable ($\Delta_\mathrm{it}$, green) components across OPSD checkpoints. Left columns: directional alignment (cosine similarity). Right columns: magnitude dominance (norm fraction). The reference component dominates in both direction and magnitude.}
\label{fig:signal-dominance}
\end{figure}

\paragraph{Results.}
The results on Qwen3-8B and R1-Distill-7B (\Figref{fig:signal-dominance}) reveal two key findings:

\subparagraph{(1) The teacher's update is dominated by the reference-induced component in both direction and magnitude.}
On both models, $\cos(\Delta_\mathrm{total}, \Delta_\mathrm{ref})$ remains high throughout training, confirming that the teacher persistently pulls the student toward the reference rather than toward question-conditioned reasoning. The norm fraction plots reinforce this: $\|\Delta_\mathrm{ref}\| / \|\Delta_\mathrm{total}\|$ consistently exceeds 1.0, meaning the reference-induced component is \emph{larger} than the total update itself. The inference-transferable component partially cancels it rather than reinforcing it.

\subparagraph{(2) The inference-transferable signal partially recovers as reference memorization saturates.}
An interesting dynamic emerges over the course of training. On Qwen3-8B, $\cos(\Delta_\mathrm{total}, \Delta_\mathrm{it})$ rises from $-$0.95 to near zero; on R1-Distill-7B, $\cos(\Delta_\mathrm{total}, \Delta_\mathrm{ref})$ rapidly climbs from 0.58 to 0.99 in the first 100 steps and then plateaus. This pattern is consistent with the reference-induced component being \emph{quickly memorized} during training: as the student absorbs the reference signal, the update along $\Delta_\mathrm{ref}$ diminishes, and the residual $\Delta_\mathrm{it}$ becomes relatively more visible. However, this partial recovery does not translate into performance improvement. As shown in \Figref{fig:opsd-collapse}, accuracy continues to decline throughout this period. The damage from early reference memorization has already destabilized the model's reasoning capability.

\vspace{0.5em}

These findings directly corroborate the performance and reflection destabilization observed in Sections~\ref{sec:revisit-fail} and \ref{sec:revisit-reflection}: \textbf{OPSD's failure on long-CoT models is caused by the dominance of reference-induced supervision ($\Delta_\mathrm{ref}$) over inference-transferable supervision ($\Delta_\mathrm{it}$) in the teacher's update signal.}

\paragraph{A natural solution emerges.}
Our decomposition identifies $\Delta_\mathrm{ref}$ as the non-transferable component of the teacher's supervision. The residual $\Delta_\mathrm{it} = \log \pi_\mathrm{T} - \log \pi_\mathrm{ref}$, obtained by subtracting this reference-induced component, can be interpreted as a conditional PMI-style residual: it measures how much the teacher's prediction changes when the question is introduced on top of the reference-conditioned context, capturing the question-conditioned correction that transfers to inference. However, $\Delta_\mathrm{it}$ is a per-token log-probability difference, not a distribution the student can directly learn from. In the following section, we show how this residual can be transformed into a well-formed target distribution, purifying the learning objective so that the student distills only the inference-transferable signal.

\section{Method}
\label{sec:method}

\subsection{PMI Target for Robust Self-Distillation}
\label{sec:pmi_target}

Our analysis in Section~\ref{sec:revisit-diagnosis} reveals that OPSD's training signal is dominated by the reference-induced component $\Delta_\mathrm{ref}$, while the question-conditioned residual
\[
    \Delta_\mathrm{it}
    =
    \log \pi_\mathrm{T}
    -
    \log \pi_\mathrm{ref}
\]
is ignored or even opposed. We interpret this residual as a conditional PMI-style quantity: it measures how much the teacher's prediction changes when the question is available in addition to the reference, capturing the inference-transferable signal we want the student to learn. However, $\Delta_\mathrm{it}$ is a log-probability difference rather than a valid probability distribution, and therefore cannot be directly used as a distillation target.

We use pointwise mutual information (PMI), a classical measure of association strength~\citep{church1990word,levy2014neural}, as the mechanism to convert this residual into a well-formed target distribution. At each token position $t$, let
\[
    P_0(v)
    =
    \pi_0(v \mid \hat{y}_{<t}, q)
\]
denote the clean base distribution conditioned on the question but not on the reference solution. We construct the PMI target by anchoring the question-conditioned residual onto this clean base distribution:
\begin{equation}
    P_{\mathrm{PMI}}(v)
    =
    \frac{1}{Z}
    P_0(v)
    \exp\!\left(
        \frac{1}{\beta}\Delta_\mathrm{it}(v)
    \right),
    \label{eq:pmi_target_distribution}
\end{equation}
where
\[
    \Delta_\mathrm{it}(v)
    =
    \log \pi_\mathrm{T}(v)
    -
    \log \pi_\mathrm{ref}(v),
\]
$\beta > 0$ controls the correction strength, and $Z$ is the normalization constant. Equivalently,
\begin{equation}
    \log P_{\mathrm{PMI}}(v)
    =
    \log P_0(v)
    +
    \frac{1}{\beta}\Delta_\mathrm{it}(v)
    -
    \log Z.
    \label{eq:pmi_target_log}
\end{equation}
This formulation starts from $P_0$, which is free of reference contamination, and adjusts it by the question-conditioned residual. When $\beta = 1$, the full correction is applied; larger $\beta$ yields a more conservative target closer to the base distribution.

\paragraph{Optimality under KL-regularized distillation.}
The PMI target in Eq.~\ref{eq:pmi_target_distribution} can be derived as the closed-form optimal distribution under a KL-regularized distillation objective. Suppose that, at a fixed token position, the transferable reward of choosing vocabulary item $v$ is the inference-transferable residual
\[
    r(v)
    =
    \Delta_\mathrm{it}(v).
\]
We seek a target distribution $P$ that maximizes this transferable reward while remaining close to the clean base distribution $P_0$:
\begin{equation}
    P^\star
    =
    \arg\max_{P \in \Delta(\mathcal{V})}
    \left[
        \mathbb{E}_{v \sim P} r(v)
        -
        \beta D_{\mathrm{KL}}(P \| P_0)
    \right].
    \label{eq:kl_regularized_objective}
\end{equation}
This objective is the token-level analogue of KL-regularized policy improvement used in RLHF and DPO-style derivations~\citep{rafailov2023direct}, with $P_0$ serving as the reference policy and $\Delta_\mathrm{it}$ serving as the implicit reward. Solving Eq.~\ref{eq:kl_regularized_objective} yields
\begin{equation}
    P^\star(v)
    =
    \frac{1}{Z}
    P_0(v)
    \exp\!\left(
        \frac{r(v)}{\beta}
    \right).
    \label{eq:optimal_policy_solution}
\end{equation}
Substituting $r(v)=\Delta_\mathrm{it}(v)$ recovers exactly the PMI target in Eq.~\ref{eq:pmi_target_distribution}. Thus, the PMI target is not a heuristic normalization of the residual; it is the optimal KL-regularized target distribution induced by the inference-transferable reward $\Delta_\mathrm{it}$ with $P_0$ as the reference policy.

For completeness, the derivation is straightforward. Introducing a Lagrange multiplier $\lambda$ for the constraint $\sum_v P(v)=1$, the Lagrangian is
\begin{equation}
    \mathcal{J}(P,\lambda)
    =
    \sum_v P(v) r(v)
    -
    \beta \sum_v P(v)\log\frac{P(v)}{P_0(v)}
    +
    \lambda\left(\sum_v P(v)-1\right).
    \label{eq:lagrangian}
\end{equation}
Setting $\partial \mathcal{J}/\partial P(v)=0$ gives
\begin{equation}
    r(v)
    -
    \beta
    \left(
        \log\frac{P(v)}{P_0(v)}
        +
        1
    \right)
    +
    \lambda
    =
    0,
\end{equation}
which implies
\begin{equation}
    P(v)
    \propto
    P_0(v)
    \exp\!\left(
        \frac{r(v)}{\beta}
    \right).
\end{equation}
After normalization and substituting $r(v)=\Delta_\mathrm{it}(v)$, we obtain Eq.~\ref{eq:pmi_target_distribution}.

\paragraph{Training objective.}
The final loss replaces the raw privileged teacher with the PMI target:
\begin{equation}
    \mathcal{L}_{\mathrm{ours}}
    =
    \frac{1}{T}
    \sum_{t=1}^{T}
    D_{\mathrm{JSD}}\!\left(
        \pi_\theta(\cdot \mid \hat{y}_{<t}, q)
        \;\|\;
        P_{\mathrm{target}}(\cdot)
    \right),
    \label{eq:final_loss}
\end{equation}
where $P_{\mathrm{target}}$ denotes the stabilized implementation of $P_{\mathrm{PMI}}$ described in Section~\ref{sec:impl}. In contrast to standard OPSD, which distills from $\pi_\mathrm{T}$ and therefore inherits the reference-induced shortcut, our objective distills only the question-conditioned correction while keeping the student anchored to its clean base reasoning prior.


\subsection{Implementation}
\label{sec:impl}

We now describe the concrete computational procedure for constructing the stabilized PMI target at each token position $t$. The procedure requires three forward passes through the same frozen base model $\pi_0$, differing only in their input prompts, plus the student's own forward pass.

\paragraph{Step 1: On-policy generation and forward passes.}
The student generates a trajectory $\hat{y} \sim \pi_\theta(\cdot \mid q)$. We then obtain three sets of logits over the full vocabulary $\mathcal{V}$ at each position $t$:
\begin{itemize}
    \item \textbf{Teacher}: $\ell_\mathrm{T}(v) = \mathrm{logit}_{\pi_0}(v \mid \hat{y}_{<t}, q, r)$ \quad (question + reference)
    \item \textbf{Reference probe}: $\ell_\mathrm{ref}(v) = \mathrm{logit}_{\pi_0}(v \mid \hat{y}_{<t}, r)$ \quad (reference only)
    \item \textbf{Base}: $\ell_0(v) = \mathrm{logit}_{\pi_0}(v \mid \hat{y}_{<t}, q)$ \quad (question only)
\end{itemize}
All three use the same frozen base model weights. The student logits
\[
    \ell_\theta(v)
    =
    \mathrm{logit}_{\pi_\theta}(v \mid \hat{y}_{<t}, q)
\]
are computed with gradients tracked.

\paragraph{Step 2: Compute the raw PMI signal.}
The inference-transferable signal is computed in log-probability space:
\begin{equation}
    \Delta_\mathrm{it}(v)
    =
    \log \pi_\mathrm{T}(v)
    -
    \log \pi_\mathrm{ref}(v),
    \quad
    \pi_\mathrm{T}
    =
    \mathrm{softmax}(\ell_\mathrm{T}),
    \quad
    \pi_\mathrm{ref}
    =
    \mathrm{softmax}(\ell_\mathrm{ref}).
    \label{eq:raw_pmi}
\end{equation}

\paragraph{Step 3: Centering.}
We subtract the vocabulary-level mean to ensure that the correction is zero-centered:
\begin{equation}
    \bar{\Delta}_\mathrm{it}(v)
    =
    \Delta_\mathrm{it}(v)
    -
    \frac{1}{|\mathcal{V}|}
    \sum_{v' \in \mathcal{V}}
    \Delta_\mathrm{it}(v').
    \label{eq:centering}
\end{equation}
Centering removes global logit shifts in the residual and makes the PMI correction numerically stable without changing its relative token-level preference structure.

\paragraph{Step 4: Soft clipping.}
We apply tanh-based soft clipping to bound extreme PMI values while preserving the direction and relative ordering of moderate values:
\begin{equation}
    \tilde{\Delta}_\mathrm{it}(v)
    =
    c \cdot
    \tanh\!\left(
        \frac{\bar{\Delta}_\mathrm{it}(v)}{c}
    \right),
    \label{eq:tanh_clip}
\end{equation}
where $c > 0$ is the clipping threshold. For $|\bar{\Delta}_\mathrm{it}(v)| \ll c$, this transformation is approximately the identity; for $|\bar{\Delta}_\mathrm{it}(v)| \gg c$, it saturates at $\pm c$. We use $c = 10$ in all experiments.

\paragraph{Step 5: Construct the stabilized PMI target.}
The practical target distribution is formed by adding the stabilized PMI signal to the base model's log-probabilities and normalizing:
\begin{equation}
    P_{\mathrm{target}}
    =
    \mathrm{softmax}\!\left(
        \log \pi_0(\cdot \mid \hat{y}_{<t}, q)
        +
        \frac{1}{\beta}
        \tilde{\Delta}_\mathrm{it}
    \right).
    \label{eq:target_construct}
\end{equation}
Equivalently,
\begin{equation}
    \log P_{\mathrm{target}}(v)
    =
    \log \pi_0(v \mid \hat{y}_{<t}, q)
    +
    \frac{1}{\beta}
    \tilde{\Delta}_\mathrm{it}(v)
    -
    \log Z_t,
    \label{eq:target_construct_log}
\end{equation}
where $Z_t$ is the token-position-specific normalization constant. We use $\beta = 1$ unless otherwise specified.

\paragraph{Step 6: Training loss.}
The student is updated by minimizing the generalized Jensen-Shannon divergence between the student distribution and the stabilized PMI target:
\begin{equation}
    \mathcal{L}
    =
    \frac{1}{T}
    \sum_{t=1}^{T}
    D_{\mathrm{JSD}}\!\left(
        \pi_\theta(\cdot \mid \hat{y}_{<t}, q)
        \;\|\;
        P_{\mathrm{target}}(\cdot)
    \right).
    \label{eq:impl_loss}
\end{equation}

\paragraph{Computational overhead.}
Compared to standard OPSD, our method requires two additional forward passes per training step: one for the reference probe and one for the base distribution. These forward passes differ only in their input prompts. Moreover, they are based on the same already generated trajectory and do not require backpropagation. As a result, our method introduces no additional trainable parameters and increases wall-clock training time by less than 10\% in our implementation. Further efficiency gains may be obtained by batching the three frozen-model forward passes together, which we leave for future work.
\section{Experiments}
\label{sec:experiments}

\subsection{Setup}
\label{sec:exp-setup}

\paragraph{Models and Datasets.} We evaluate on four long-CoT reasoning models spanning different architectures and scales: Qwen3-8B, Qwen3-4B~\citep{yang2025qwen3}, DeepSeek-R1-Distill-Qwen-7B~\citep{guo2025deepseek} (abbreviated R1-Distill-7B), and OLMo-7B-Thinking~\citep{olmo2025olmo}. We use two training datasets with reference solutions as privileged information: \textbf{DASD-10K}, a 10K-sample subset from DASD~\citep{yan2026distributionalignedsequencedistillationsuperior}, and \textbf{Math-CoT-20K}~\citep{ren2026rethinkinggeneralizationreasoningsft}, a 20K-sample collection of competition-level math problems with detailed chain-of-thought solutions.

\paragraph{Baselines and Evaluation.} We compare three configurations: (1)~\textbf{Base}: the base model without any distillation, (2)~\textbf{OPSD-Standard}: standard on-policy self-distillation~\citep{zhao2026self} with JSD loss, and (3)~\textbf{OPSD-PMI (Ours)}: on-policy self-distillation with our PMI target (Section~\ref{sec:pmi_target}). Following OPSD, we evaluate checkpoints every 50 steps up to 200 steps and report best accuracy on three challenging mathematical reasoning benchmarks: \textbf{AIME 2024}, \textbf{AIME 2025}, and \textbf{HMMT 2025}. All the scores are 12-run averages.

\paragraph{Implementation details.} Unless otherwise specified, we follow the training configuration of OPSD. All methods use LoRA~\citep{hu2022lora} adaptation with rank 64. Training uses a learning rate of 5e-6, a batch size of 32, and gradient checkpointing. On-policy trajectories are generated using vLLM~\citep{kwon2023efficient}. The PMI target uses $\beta = 1$, tanh soft clipping with $c = 10$, and centering. Maximum completion length is 1024 tokens during training. We do not apply any post-hoc token-level loss clipping, as the tanh soft clipping in the PMI target construction (Eq.~\ref{eq:tanh_clip}) already bounds extreme values.

\subsection{Main Results}
\label{sec:exp-main}

\begin{table}[t]
\centering
\caption{Main results on mathematical reasoning benchmarks. \textbf{Bold} indicates the best result per model-dataset combination. OPSD-Standard uses the raw teacher distribution as the distillation target; OPSD-PMI (Ours) replaces it with our purified PMI target. OPSD-PMI consistently improves over both the base model and OPSD-Standard across all four models and both training datasets.}
\label{tab:main-results}
\resizebox{\textwidth}{!}{
\begin{tabular}{ll cccc cccc}
\toprule
& & \multicolumn{4}{c}{\textbf{DASD-10K}} & \multicolumn{4}{c}{\textbf{Math-CoT-20K}} \\
\cmidrule(lr){3-6} \cmidrule(lr){7-10}
\textbf{Model} & \textbf{Method} & AIME24 & AIME25 & HMMT25 & Avg. & AIME24 & AIME25 & HMMT25 & Avg. \\
\midrule
\multirow{3}{*}{Qwen3-8B}
& Base & 75.8 & 65.6 & 43.9 & 61.8 & 75.8 & 65.6 & 43.9 & 61.8 \\
& OPSD-Standard & 75.4 & 65.2 & 42.2 & 60.9 & 75.8 & 66.7 & 44.4 & 62.3 \\
& OPSD-PMI (Ours) & \textbf{79.4} & \textbf{71.9} & \textbf{46.7} & \textbf{66.0} & \textbf{77.1} & \textbf{70.8} & \textbf{47.5} & \textbf{65.1} \\
\midrule
\multirow{3}{*}{Qwen3-4B}
& Base & 74.9 & 66.4 & 42.2 & 61.2 & 74.9 & 66.4 & 42.2 & 61.2 \\
& OPSD-Standard & 74.2 & 65.2 & 42.2 & 60.5 & 73.3 & 64.2 & 40.8 & 59.4 \\
& OPSD-PMI (Ours) & \textbf{76.3} & \textbf{68.3} & \textbf{46.4} & \textbf{63.7} & \textbf{76.1} & \textbf{67.5} & \textbf{44.4} & \textbf{62.7} \\
\midrule
\multirow{3}{*}{R1-Distill-7B}
& Base & 52.0 & 39.6 & 24.4 & 38.7 & 52.0 & 39.6 & 24.4 & 38.7 \\
& OPSD-Standard & 51.9 & 39.2 & 24.4 & 38.5 & 52.2 & 38.1 & 24.4 & 38.2 \\
& OPSD-PMI (Ours) & \textbf{54.0} & \textbf{43.1} & \textbf{25.3} & \textbf{40.8} & \textbf{55.3} & \textbf{41.1} & \textbf{26.1} & \textbf{40.8} \\
\midrule
\multirow{3}{*}{OLMo-7B}
& Base & 71.9 & 66.7 & 45.2 & 61.3 & 71.9 & 66.7 & 45.2 & 61.3 \\
& OPSD-Standard & 68.1 & 66.4 & 45.0 & 59.8 & 69.7 & 65.6 & 41.7 & 59.0 \\
& OPSD-PMI (Ours) & \textbf{74.7} & \textbf{68.9} & \textbf{46.1} & \textbf{63.2} & \textbf{73.3} & \textbf{70.3} & \textbf{46.4} & \textbf{63.3} \\
\bottomrule
\end{tabular}
}
\end{table}

Table~\ref{tab:main-results} presents the main results across four models and two training datasets. We highlight three key findings:

\textbf{(1) OPSD-Standard fails on long-CoT models.}
Consistent with our analysis in Section~\ref{sec:revisit}, OPSD-Standard provides negligible or negative gains across all settings. On DASD-10K, OPSD-Standard degrades performance on three of four models, with only R1-Distill-7B showing near-zero change. On Math-CoT-20K, OPSD-Standard provides a marginal gain on Qwen3-8B but degrades the other three models, with OLMo-7B suffering the largest drop. These results confirm that the reference-induced supervision dominance identified in Section~\ref{sec:revisit-diagnosis} translates directly into practical failure.

\textbf{(2) OPSD-PMI consistently improves over the base model.}
In contrast to OPSD-Standard's failure, OPSD-PMI achieves gains on \emph{\textbf{every}} model-dataset combination. These results demonstrate that the inference-transferable signal, once isolated via the reference-only teacher and transformed into a learnable target via PMI, provides genuinely useful supervision for long-CoT models.

\textbf{(3) The gap between OPSD-Standard and OPSD-PMI is large and consistent.}
Comparing the two methods directly, OPSD-PMI outperforms OPSD-Standard on both DASD-10K and Math-CoT-20K. This gap is remarkably consistent across models and datasets, suggesting that the reference-induced noise is a \textbf{\emph{universal}} bottleneck in OPSD-Standard for long-CoT models, and that OPSD-PMI addresses it effectively regardless of the specific model architecture or training data.

\begin{figure}[t]
\centering
\subfigure[Qwen3-8B]{
    \includegraphics[width=0.23\textwidth,keepaspectratio]{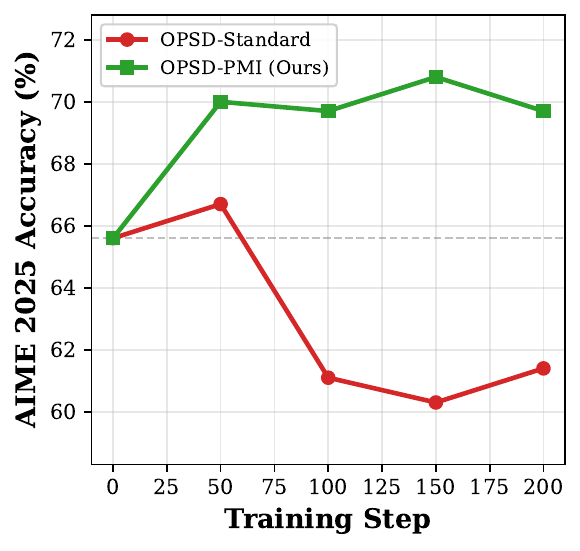}
    \label{fig:dyn-8b}
}
\subfigure[Qwen3-4B]{
    \includegraphics[width=0.23\textwidth,keepaspectratio]{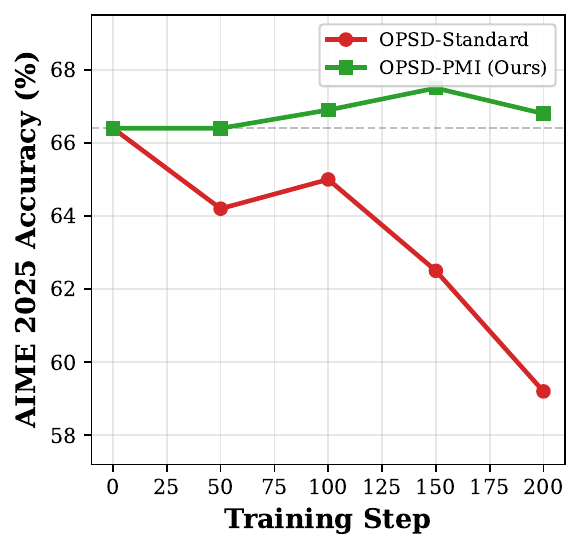}
    \label{fig:dyn-4b}
}
\subfigure[R1-Distill-7B]{
    \includegraphics[width=0.23\textwidth,keepaspectratio]{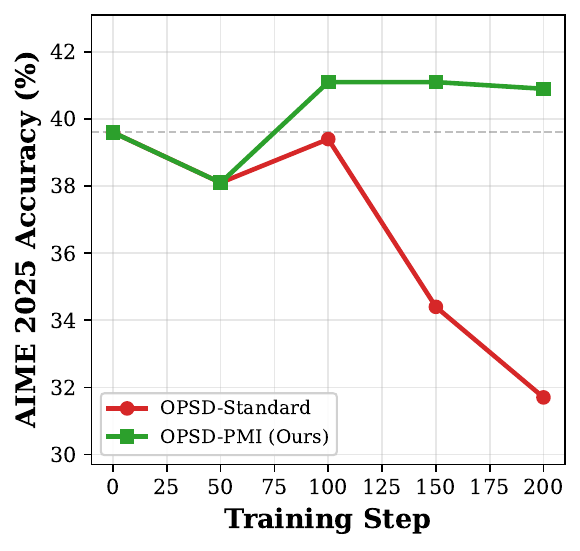}
    \label{fig:dyn-r1}
}
\subfigure[OLMo-7B]{
    \includegraphics[width=0.23\textwidth,keepaspectratio]{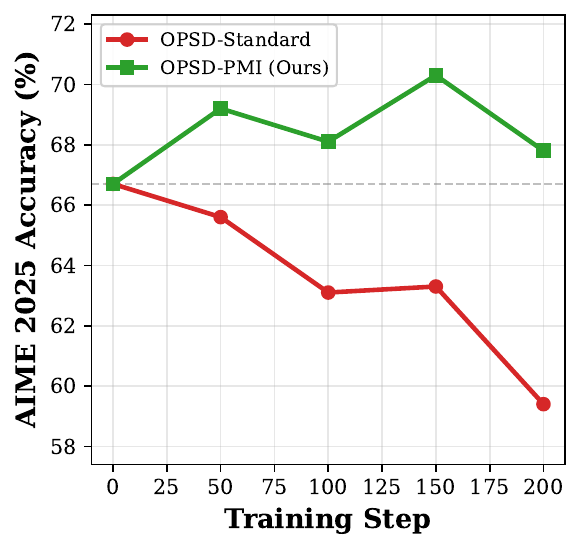}
    \label{fig:dyn-olmo}
}
\caption{Training dynamics on Math-CoT-20K (AIME 2025). OPSD-Standard (red) uses the raw teacher as the distillation target and degrades progressively; OPSD-PMI (Ours, green) replaces it with our purified PMI target and remains stably above the baseline (dashed) throughout training.}
\label{fig:training-dynamics}
\end{figure}

\paragraph{Training dynamics.}
While the main results table reports best-checkpoint performance, \Figref{fig:training-dynamics} reveals the full training trajectory, directly contrasting with the OPSD-Standard collapse shown in \Figref{fig:opsd-collapse}. Two key observations emerge. First, \textbf{OPSD-PMI improves and remains stable}, whereas OPSD-Standard peaks briefly (if at all) before steadily declining. Second, \textbf{OPSD-PMI is robust to checkpoint selection}: the variance across checkpoints is small, meaning practitioners do not need careful early stopping to avoid catastrophic degradation. This is a practical advantage over OPSD-Standard, where later checkpoints consistently perform worse.

\subsection{Preservation of Reflective Reasoning}
\label{sec:exp-epistemic}

A central claim of our work is that OPSD-PMI preserves the reflective reasoning capability that OPSD-Standard destabilizes (Section~\ref{sec:revisit-reflection}). To verify this, we revisit the epistemic marker analysis on Qwen3-8B and R1-Distill-7B, now comparing the two methods side by side.

\begin{figure}[t]
\centering
\subfigure[Qwen3-8B: epistemic count]{
    \includegraphics[width=0.23\textwidth,keepaspectratio]{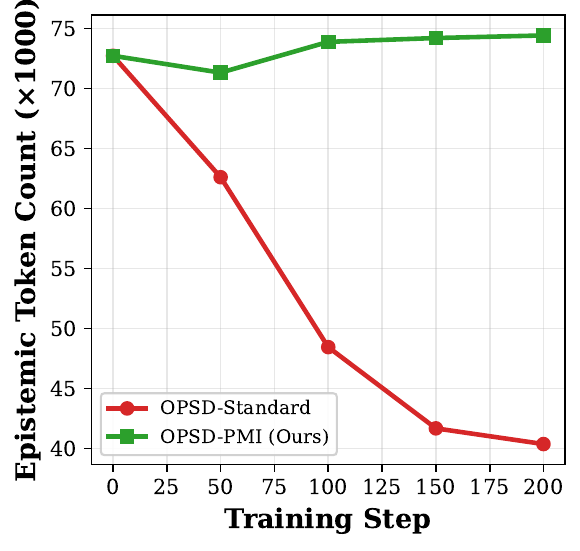}
    \label{fig:exp-qwen-traj}
}
\subfigure[Qwen3-8B: marker distribution]{
    \includegraphics[width=0.23\textwidth,keepaspectratio]{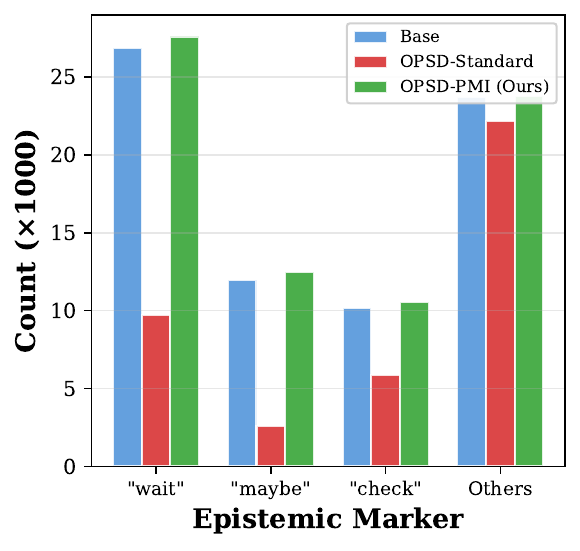}
    \label{fig:exp-qwen-markers}
}
\subfigure[R1-7B: epistemic count]{
    \includegraphics[width=0.23\textwidth,keepaspectratio]{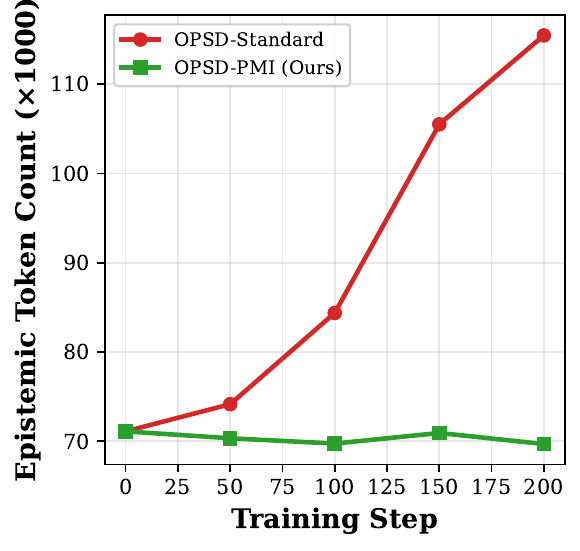}
    \label{fig:exp-r1-traj}
}
\subfigure[R1-7B: marker distribution]{
    \includegraphics[width=0.23\textwidth,keepaspectratio]{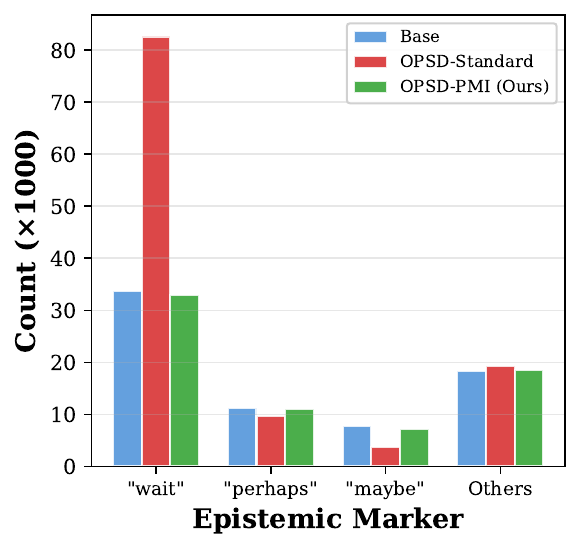}
    \label{fig:exp-r1-markers}
}
\caption{Epistemic marker analysis on Math-CoT-20K. OPSD-Standard uses the raw teacher distribution; OPSD-PMI (Ours) uses our purified PMI target. (a, c)~Total epistemic token count across training steps: OPSD-Standard causes pathological collapse or explosion, while OPSD-PMI remains stable at baseline levels. (b, d)~Per-marker distribution at step 200: OPSD-Standard distorts the distribution, while OPSD-PMI closely preserves the base model's natural epistemic behavior.}
\label{fig:exp-epistemic}
\end{figure}

\paragraph{Epistemic marker trajectories.}
\Figref{fig:exp-epistemic} (a, c) shows the total epistemic token count across training checkpoints. Under OPSD-Standard, Qwen3-8B's epistemic count collapses from 73K to 40K, while R1-Distill-7B's explodes from 71K to 115K. In stark contrast, OPSD-PMI maintains the epistemic count at near-baseline levels throughout training on both models: Qwen3-8B and R1-Distill-7B remain within the 70K range. The stability is remarkable: OPSD-PMI trains the model to improve its reasoning accuracy without disrupting its reflective behavior.

\paragraph{Marker distribution preservation.}
\Figref{fig:exp-epistemic} (b, d) compares the per-marker distribution at the final checkpoint. On Qwen3-8B, OPSD-Standard uniformly suppresses all marker types (``\textit{wait}'' drops from 27K to 10K, ``\textit{maybe}'' from 12K to 3K), while OPSD-PMI closely preserves the base distribution across all markers. On R1-Distill-7B, OPSD-Standard concentrates the increase almost entirely on ``\textit{wait}'' (34K $\to$ 83K) while other markers remain stable or decline, a degenerate pattern suggesting repetitive rather than genuine deliberation. OPSD-PMI maintains a balanced distribution nearly identical to the base model.

These results are consistent with our diagnosis that the reference-induced supervision identified in Section~\ref{sec:revisit-diagnosis} is the main driver of epistemic destabilization, and OPSD-PMI, by filtering out this component, preserves the model's natural reflective reasoning while still enabling effective knowledge transfer.

\subsection{Ablation Studies}
\label{sec:ablation}

We ablate the two key hyperparameters of the PMI target, namely the soft clipping threshold $c$ and the correction strength $\beta$, on Qwen3-8B and R1-Distill-7B using Math-CoT-20K. We evaluate on AIME 2024 and AIME 2025 across checkpoints 50 to 200.

\begin{figure}[t]
\centering
\subfigure[Qwen3-8B (AIME24)]{
    \includegraphics[width=0.23\textwidth,keepaspectratio]{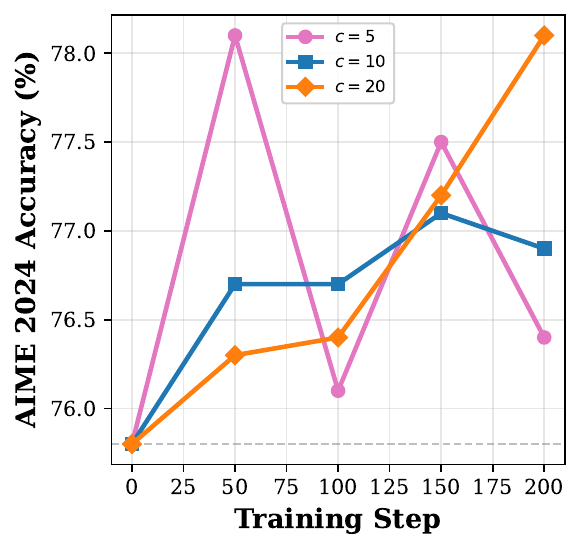}
    \label{fig:abl-qwen-clip-24}
}
\subfigure[Qwen3-8B (AIME25)]{
    \includegraphics[width=0.23\textwidth,keepaspectratio]{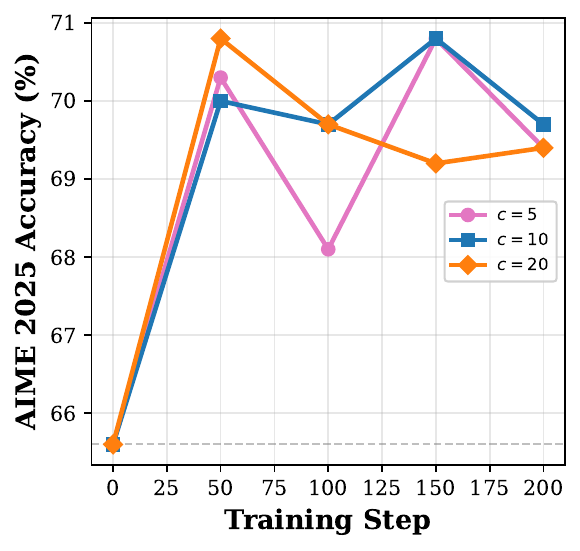}
    \label{fig:abl-qwen-clip-25}
}
\subfigure[R1-7B (AIME24)]{
    \includegraphics[width=0.23\textwidth,keepaspectratio]{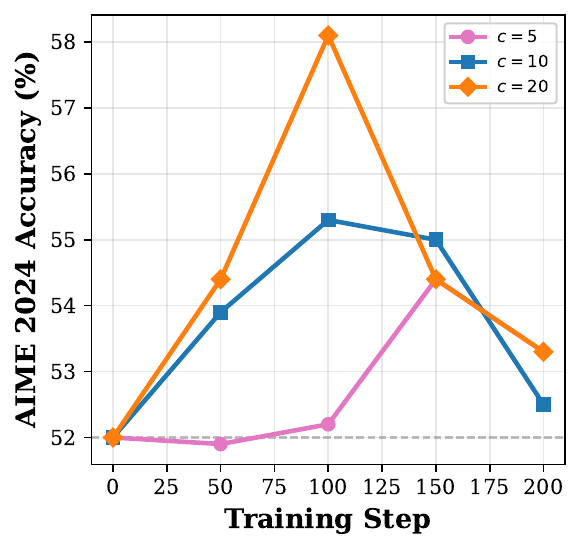}
    \label{fig:abl-r1-clip-24}
}
\subfigure[R1-7B (AIME25)]{
    \includegraphics[width=0.23\textwidth,keepaspectratio]{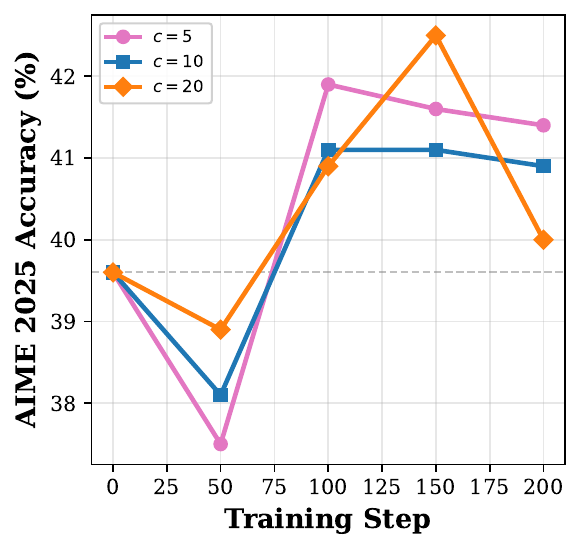}
    \label{fig:abl-r1-clip-25}
}
\caption{Ablation on the soft clipping threshold $c$ ($\beta = 1$ fixed). All settings consistently improve over the baseline and show similar trajectories, confirming robustness to this hyperparameter.}
\label{fig:ablation-clip}
\end{figure}

\paragraph{Soft clipping threshold $c$.}
\Figref{fig:ablation-clip} compares $c \in \{5, 10, 20\}$ with $\beta = 1$ fixed. All three settings consistently improve over the baseline and substantially outperform OPSD-Standard across both models and benchmarks. $c = 20$ exhibits training trajectories closely matching $c = 10$, as most PMI values naturally fall well within this range. $c = 5$ introduces slightly more volatile trajectories, it may be because extreme PMI values are more aggressively compressed by the tanh nonlinearity at this threshold, which may cause minor distortion. However, it does not much lower the performance ceiling. Overall, the method is robust to the choice of $c$: the tanh soft clipping serves as a safety net for numerical stability rather than a performance-critical component.

\begin{figure}[t]
\centering
\subfigure[Qwen3-8B (AIME24)]{
    \includegraphics[width=0.23\textwidth,keepaspectratio]{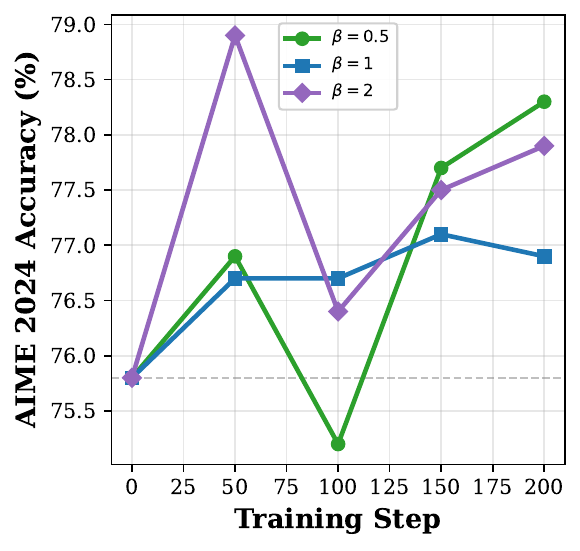}
    \label{fig:abl-qwen-beta-24}
}
\subfigure[Qwen3-8B (AIME25)]{
    \includegraphics[width=0.23\textwidth,keepaspectratio]{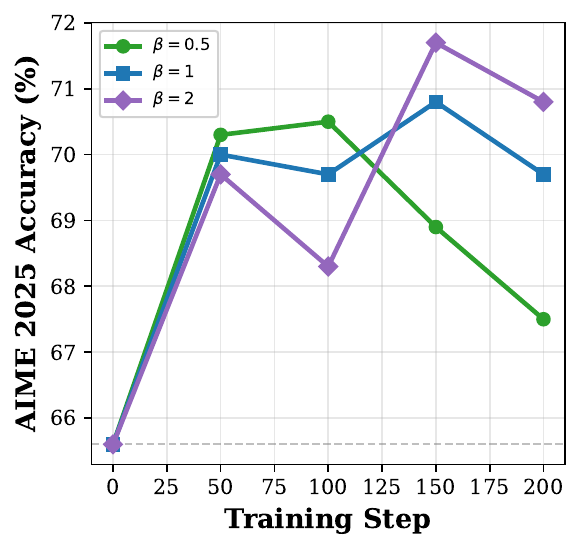}
    \label{fig:abl-qwen-beta-25}
}
\subfigure[R1-7B (AIME24)]{
    \includegraphics[width=0.23\textwidth,keepaspectratio]{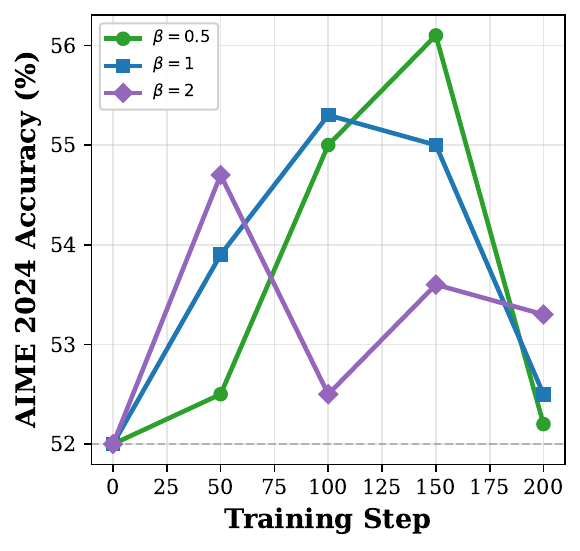}
    \label{fig:abl-r1-beta-24}
}
\subfigure[R1-7B (AIME25)]{
    \includegraphics[width=0.23\textwidth,keepaspectratio]{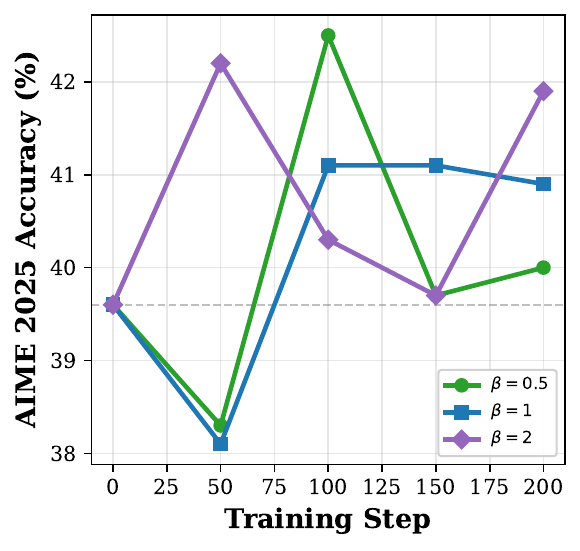}
    \label{fig:abl-r1-beta-25}
}
\caption{Ablation on the correction strength $\beta$ ($c = 10$ fixed). $\beta = 0.5$ produces volatile trajectories; $\beta = 2$ yields smoother curves with occasionally higher peaks; $\beta = 1$ balances stability and performance.}
\label{fig:ablation-beta}
\end{figure}

\paragraph{Correction strength $\beta$.}
\Figref{fig:ablation-beta} compares $\beta \in \{0.5, 1, 2\}$ with $c = 10$ fixed. All three settings improve over both the baseline and OPSD-Standard, but exhibit distinct training dynamics. The performance curves differ across settings: different $\beta$ values achieve the best result at different checkpoints and benchmarks, with no single value uniformly dominating. This is expected, since $\beta$ controls the trade-off between the strength of the question-conditioned correction and proximity to the base distribution. Importantly, all settings consistently outperform both the base model and OPSD-Standard, demonstrating that OPSD-PMI is robust to the choice of $\beta$. For simplicity and to avoid additional tuning, we use $\beta = 1$ in all main experiments.

\section{Related Work}
\label{sec:related}

\paragraph{Long chain-of-thought reasoning.}
Chain-of-thought (CoT) prompting~\citep{wei2022chain} demonstrated that eliciting intermediate reasoning steps substantially improves LLM performance on complex tasks. This insight has been scaled to long chain-of-thought (long-CoT) reasoning, where models engage in extended, reflective thinking processes including self-correction, backtracking, and uncertainty externalization~\citep{jaech2024openai,guo2025deepseek,team2024qwq}. Models such as OpenAI o1~\citep{jaech2024openai}, DeepSeek-R1~\citep{guo2025deepseek}, and QwQ~\citep{team2024qwq} have shown that scaling test-time compute through longer reasoning chains can be more effective than scaling model parameters~\citep{snell2024scaling}. A key characteristic of long-CoT models is their use of epistemic markers, such as ``\textit{Wait}'', ``\textit{Perhaps}'', and ``\textit{Let me reconsider}'', that externalize uncertainty and enable reflective deliberation~\citep{gandhi2024stream,li2025feature}. Distilling long-CoT capabilities into smaller models has attracted significant attention. Supervised approaches include fine-tuning on curated long-CoT traces~\citep{muennighoff2025s1,ye2025limo,xu2025redstar,wen2025light}, structured decomposition of reasoning chains~\citep{luo2025deconstructing,chen2025unveiling}, data-efficient frameworks~\citep{wu2025beyond}, progressive training through the ``valley'' of capability~\citep{luo2025through}, chunk-wise distillation~\citep{chen2025skip}, and long-CoT compression~\citep{wang2025r1}. However, these off-policy approaches suffer from train-test distribution mismatch, as the student trains on teacher-generated trajectories rather than its own. Our work addresses this limitation through on-policy distillation, while identifying and solving a fundamental failure mode specific to long-CoT models.

\paragraph{On-policy self-distillation.}
Knowledge distillation~\citep{hinton2015distilling} transfers knowledge from a teacher to a student via soft label matching. On-Policy Self-Distillation (OPSD)~\citep{agarwal2024onpolicy} extends this paradigm by having the student generate its own trajectories and receive token-level feedback from a privileged teacher with access to reference solutions, directly addressing the distribution mismatch problem inherent in off-policy distillation~\citep{yu2024distilling}. Several concurrent works have extended OPSD: \citet{zhao2026self} proposed Self-Distilled Reasoner for general LLM reasoning, \citet{he2026self} introduced Self-Distillation Zero combining self-revision with binary rewards to provide dense supervision, and \citet{cui2026brief} provided a comprehensive overview of the OPSD landscape. However, recent investigations have revealed critical limitations when applying OPSD to thinking models. \citet{kaur2026rethinking} showed that standard OPSD provides limited gains on models with strong reasoning capabilities, and \citet{kim2026does} investigated why self-distillation sometimes degrades reasoning, identifying the destabilization of epistemic markers as a key symptom. Our work makes three advances beyond these findings: (1)~we provide a \emph{mechanistic diagnosis} through a decomposition of the teacher's supervision signal, showing that a reference-only teacher can isolate the non-transferable component and reveal that the reference-induced signal dominates while the inference-transferable residual is ignored or opposed; (2)~we propose the \emph{PMI target}, which leverages pointwise mutual information as the mechanism to transform this residual into a purified target distribution; and (3)~we demonstrate that this approach consistently improves performance while preserving the reflective reasoning capability across multiple long-CoT models.

\section{Conclusion}
\label{sec:conclusion}

We investigated why on-policy self-distillation (OPSD) fails on long-CoT reasoning models. Through a decomposition of the teacher's supervision signal, we showed that the reference-induced component dominates both the direction and magnitude of the training signal, while the question-conditioned, inference-transferable component is suppressed or opposed. This explains the observed performance degradation and epistemic marker destabilization: OPSD drives rote memorization of the reference rather than transferable reasoning improvement.

Based on this diagnosis, we proposed a two-step solution: first, constructing a reference-only teacher to isolate the non-transferable component and reveal the question-conditioned residual; second, using pointwise mutual information (PMI) as the mechanism to anchor this residual onto the base distribution, producing a purified PMI target that the student can directly distill from. The method integrates seamlessly into the standard OPSD framework with minimal overhead. Across four long-CoT models and two training datasets, our approach consistently improves over both the base model and standard OPSD, while preserving the models' reflective reasoning capability throughout training. Our ablation studies also confirm robustness to key hyperparameters.



\newpage
\bibliographystyle{plainnat}
\bibliography{ref}
\appendix

\end{document}